\documentclass[10pt,twocolumn,letterpaper]{article}

 \usepackage[pagenumbers]{cvpr} % To force page numbers, e.g. for an arXiv version
\usepackage{multirow}

\usepackage[dvipsnames]{xcolor}

\definecolor{cvprblue}{rgb}{0.21,0.49,0.74}
\usepackage[pagebackref,breaklinks,colorlinks,citecolor=cvprblue]{hyperref}

\def\paperID{14614} % *** Enter the Paper ID here
\def\confName{ICCV}
\def\confYear{2025}

\title{Contrast: A Hybrid Architecture of Transformers and State Space Models for Low-Level Vision}

\author{Aman Urumbekov\\
Kyrgyz State Technical University\\
{\tt\small amanurumbekov@gmail.com}
\and
Zheng Chen\\
Shanghai Jiao Tong University\\
{\tt\small zhengchen.cse@gmail.com}
}

\begin{document}
\maketitle
\begin{abstract} 
Transformers have emerged as a leading approach for image super-resolution (SR) tasks due to their strong global context modeling capabilities. However, their quadratic computational complexity necessitates window-based attention, which restricts the receptive field and hampers broader context expansion. Recently, Mamba has been introduced as a promising alternative with linear complexity, enabling it to dispense with window-based mechanisms while maintaining a large receptive field. Nevertheless, Mamba struggles with long-range dependencies when high pixel-level precision is required—an essential aspect of SR—because its hidden state mechanism can only approximate the vast amount of stored context, causing inaccuracies not present in transformers. To address these limitations, we propose \textbf{Contrast}, a hybrid SR model that integrates \textbf{Con}volutional, \textbf{Tra}nsformer, and \textbf{St}ate Space components. By uniting transformer and state space modules, \textbf{Contrast} harnesses the strengths of each method, enhancing global context modeling and improving pixel-level accuracy. We show that this fusion mitigates the inherent shortcomings of both approaches, thereby achieving superior performance on image super-resolution tasks.
\end{abstract}    
\section{Introduction}

\begin{figure}[t]
    \centering
    \begin{minipage}[t]{0.45\textwidth}
        \includegraphics[width=\textwidth]{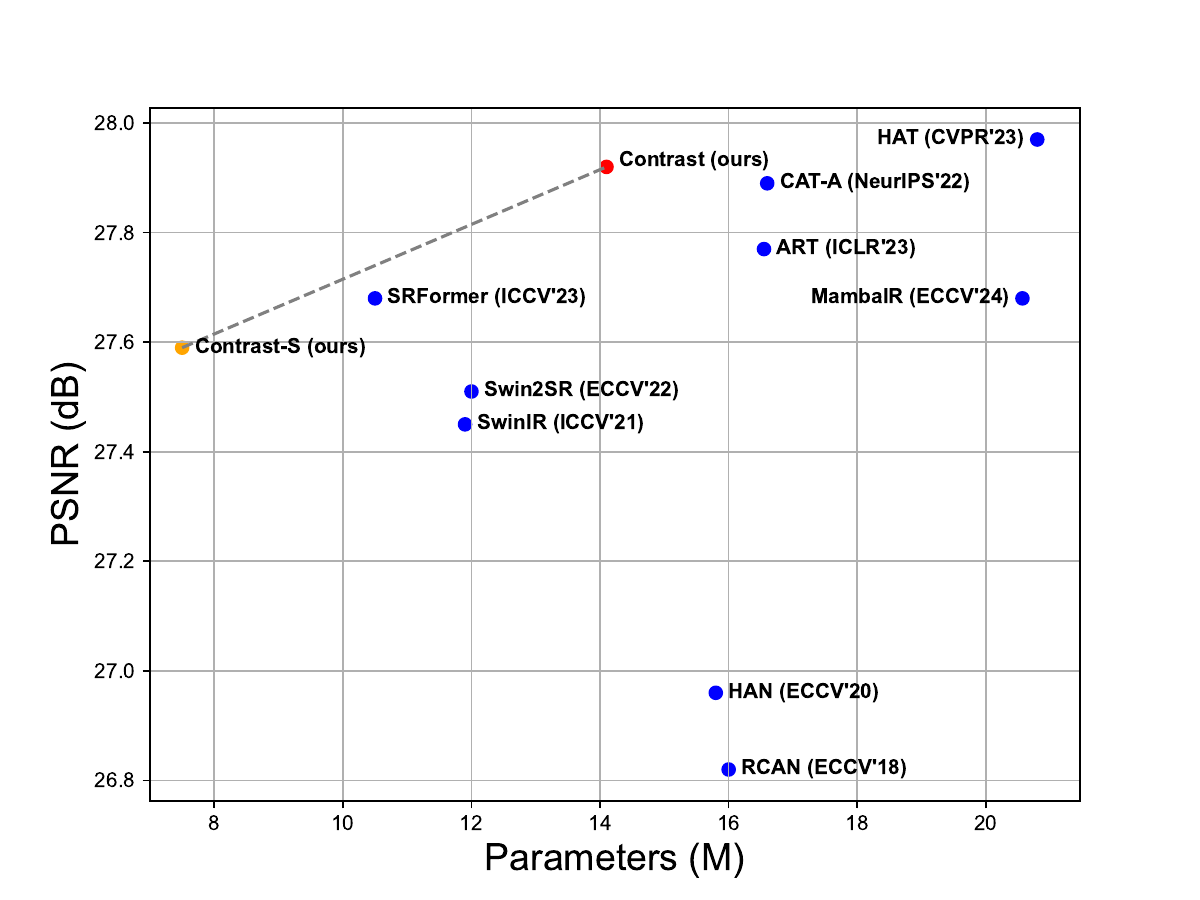}
        \caption{Model comparison on the Urban100 dataset for $\times$4 SR. The plot shows the trade-off between PSNR and parameter count, where higher PSNR values and fewer parameters (upper-left) indicate better performance. \textbf{Contrast} demonstrates superior effectiveness on the Urban100 benchmark.}
        \label{first_fig}
    \end{minipage}
\end{figure}

Single image super-resolution (SR) aims to reconstruct a high-resolution (HR) image from a low-resolution (LR) input and serves as a foundational task in low-level vision. Because multiple valid HR outputs can correspond to the same LR input, SR is inherently ill-posed. Over the past decade, various methods have been explored to enhance SR performance, balancing computational efficiency with the capacity to capture fine-grained details.

\textbf{Convolutional Neural Networks (CNNs)}~\cite{he2016deep, huang2017densely, simonyan2014very} were among the earliest architectures employed for SR~\cite{dong2015imagesuperresolutionusingdeep, dai2019second, mei2020image, zhang2018image}. While they effectively model local structures, their reliance on localized convolutions hinders their ability to capture long-range dependencies—an essential factor for contextual understanding in complex images. This limitation prompted investigations into alternative paradigms better suited for modeling global context.

\textbf{Transformers}~\cite{vaswani2017attention}, initially introduced for natural language processing (NLP), have demonstrated strong potential in computer vision, especially in high-level tasks~\cite{dosovitskiy2020image, touvron2021training, tu2022maxvit}. Their self-attention (SA) mechanism directly models long-range dependencies, helping mitigate the locality constraints inherent to CNNs. However, transformers exhibit quadratic computational complexity with increasing image resolution, which becomes prohibitive for high-resolution SR. To manage this, SR models typically adopt window-based self-attention~\cite{liu2021swin, liang2021swinir}, limiting attention calculations to local windows to reduce complexity. Although this strategy addresses the computational challenge, it constrains the receptive field, thus reducing access to broader contextual information needed for high-quality SR.

Recently, the \textbf{Mamba} architecture~\cite{mamba, mamba2} has emerged as an alternative emphasizing efficient global modeling with linear complexity. Mamba can extend its receptive field across the entire image without resorting to window mechanisms, circumventing the restricted receptive field encountered in window-based attention. This feature is particularly advantageous for SR tasks, where capturing global context is pivotal. Moreover, Mamba’s efficiency does not rely on partitioning the input into windows, allowing it to preserve a global perspective by default.

However, Mamba encounters challenges in tasks requiring high pixel-level precision, such as SR. Although its hidden state mechanism effectively stores substantial context, this representation is approximate, potentially causing inaccuracies in long-context modeling and ultimately affecting image quality. In contrast, transformers achieve more accurate results in such scenarios due to their precise attention patterns.

In this paper, we propose \textbf{Contrast}, a hybrid SR model that combines the strengths of both transformers and Mamba to overcome their respective limitations. Specifically, \textbf{Contrast} integrates \textbf{Con}volutional, \textbf{Tra}nsformer, and \textbf{St}ate Space components, leveraging Mamba’s efficient global receptive field alongside the pixel-level precision of transformers. By fusing these complementary capabilities, Contrast provides robust modeling of both local and global context, offering a balanced approach that neither method can achieve individually.

As illustrated in Figure~\ref{first_fig}, Contrast achieves superior results on the Urban100~\cite{huang2015single} dataset, which features high-resolution images with repetitive structures—a scenario well-suited for models with an extensive receptive field. Specifically, Contrast reaches a PSNR of 27.92 with only 14.1 million parameters, outperforming MambaIR (ECCV’24)~\cite{guo2024mambair}, which achieves 27.68 PSNR using 20.6 million parameters, and closely matching HAT (CVPR’23)~\cite{chen2023activating}, which attains 27.97 PSNR but with 20.8 million parameters. Contrast’s parameter efficiency stems from its hybrid design, employing one transformer block for every six Mamba blocks, thereby providing high-quality reconstruction with fewer parameters than competing methods.

In summary, window mechanisms in transformers mitigate the challenge of quadratic complexity but inevitably restrict the model’s receptive field, making it difficult to capture extensive context. Meanwhile, Mamba’s linear complexity obviates window-based operations, naturally maintaining a global receptive field. By integrating these two paradigms, Contrast capitalizes on their complementary benefits, yielding superior performance in image super-resolution tasks.

\begin{figure*}[t]
    \centering
    \includegraphics[width=1\textwidth]{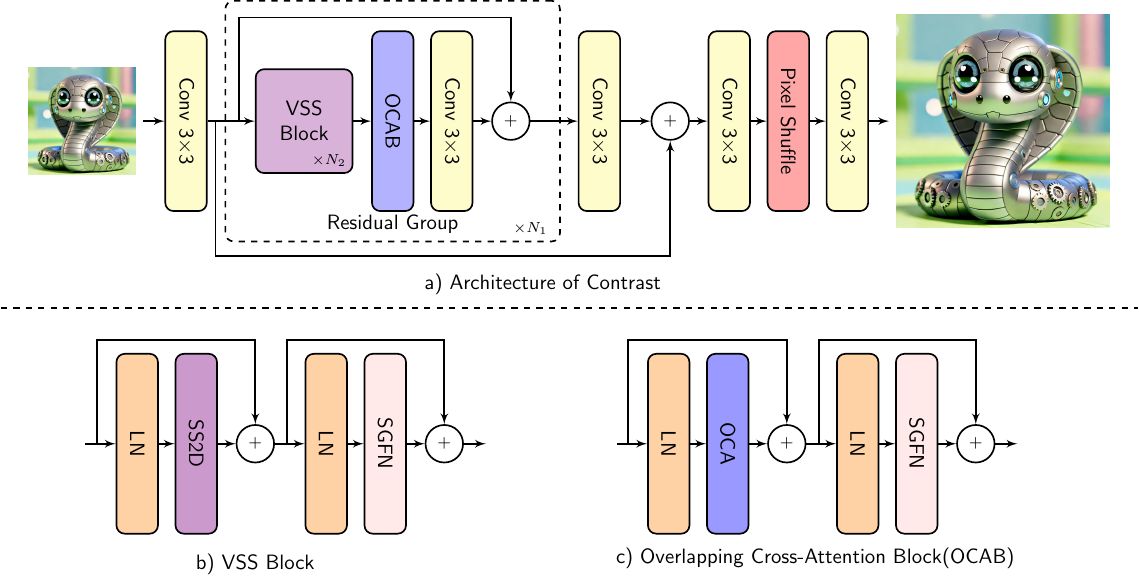}
    \vspace{-6mm}
    \caption{Overall architecture of the proposed Contrast model. (a) The high-level Contrast framework. (b) Detailed structure of the VSS Block. (c) Detailed structure of the Overlapping Cross-Attention Block (OCAB).}
    \label{fig:contrast_arch}
    \vspace{-2mm}
\end{figure*}
\section{Related Work}

\textbf{Image Super-Resolution.} Deep learning has revolutionized image super-resolution (SR), with Convolutional Neural Networks (CNNs) initially setting the benchmark for performance. Pioneering work like SRCNN~\cite{dong2015imagesuperresolutionusingdeep} introduced CNNs to SR, achieving significant improvements over traditional methods. Advanced architectures such as RCAN~\cite{zhang2018image} utilized deep residual networks exceeding 400 layers to enhance feature extraction. Additionally, attention mechanisms~\cite{zhang2018image, wang2018esrgan} were integrated to focus on important spatial and channel-wise information. Despite these advancements, CNNs inherently struggle to capture long-range dependencies due to the local nature of convolution operations, limiting their ability to model global context effectively.

\textbf{Vision Transformers in Low-Level Vision.} Transformers~\cite{vaswani2017attention}, originally designed for natural language processing, have shown great promise in computer vision tasks~\cite{dosovitskiy2020image, touvron2021training, tu2022maxvit} due to their self-attention mechanism, which can model long-range dependencies. In SR, Transformers help overcome the limitations of CNNs by capturing global context. However, the quadratic computational complexity of self-attention with respect to input size poses significant challenges for high-resolution images typical in SR tasks. To alleviate this, models like SwinIR~\cite{liu2021swin, liang2021swinir} employ window-based self-attention, where attention is calculated within local windows to reduce computational costs. While this approach effectively mitigates quadratic complexity, it introduces a limited receptive field that is difficult to expand, constraining the model's ability to capture broader image context essential for high-quality SR outcomes.

\textbf{State Space Models.} State space models (SSMs) have recently emerged as a promising alternative for modeling long-range dependencies with linear computational complexity. Architectures like Mamba~\cite{mamba, mamba2} exploit SSMs to efficiently capture global context, inherently providing a global receptive field. This makes them particularly attractive for SR tasks, where understanding the entire image context is crucial. However, SSM-based models like Mamba face challenges when precise pixel-level accuracy is required. The hidden state mechanism compresses context information in an approximate manner, which can lead to inaccuracies in modeling long-context dependencies. Additionally, their sequential scanning mechanism makes it difficult to capture diagonal dependencies, which are important for reconstructing detailed spatial patterns.

\textbf{Hybrid Models.} Numerous architectures have explored hybrid models that integrate Transformers with SSMs like Mamba, often employing full attention mechanisms in the Transformer components~\cite{waleffe2024empiricalstudymambabasedlanguage}. However, few studies have delved into providing a comprehensive explanation for the effectiveness of such integrations.

\section{Methodology}

\subsection{Motivation}

Recent super-resolution (SR) approaches frequently employ Transformer-based architectures due to their precise attention mechanisms and strong modeling capacity. However, because of the quadratic complexity in self-attention, these models are often forced to rely on window-based operations to remain computationally tractable, thereby restricting their effective receptive field. This limitation can be problematic for SR tasks involving high-resolution images with intricate global structures.

On the other hand, Mamba architectures feature linear complexity and inherently capture global context without resorting to window partitioning. Their hidden state mechanism, however, can introduce approximation errors that become critical for pixel-level accuracy—an essential requirement in SR. To address these contrasting limitations, our proposed model integrates both Transformer and Mamba components: the Mamba blocks supply a global receptive field free of window constraints, while Transformer attention layers refine these global features with high precision. Furthermore, we replace the standard MLP modules with convolution-based SGFNs, facilitating better local context modeling. As a result, our hybrid approach offers both robust global context handling and pixel-level detail recovery, simultaneously addressing the shortcomings of purely Transformer-based or purely Mamba-based SR solutions.

\subsection{Architecture}

The Contrast model is designed as a hybrid framework with three primary modules: shallow feature extraction, deep feature extraction, and image reconstruction, as illustrated in Fig.~\ref{fig:contrast_arch}. This architecture balances computational efficiency and performance in super-resolution (SR) tasks by combining Visual State Space (VSS) Blocks\cite{liu2024vmamba}, Overlapping Cross-Attention Blocks (OCAB)\cite{chen2023activating}, and Spatial Gated Feed-Forward Networks (SGFN)\cite{chen2023dual}. Each module is tailored to extract and refine features at multiple levels, progressively enhancing the image from low resolution to high resolution.

\subsubsection{Shallow Feature Extraction}

Given a low-resolution (LR) input image $I_{LR} \in \mathbb{R}^{H \times W \times 3}$, the shallow feature extraction module first applies a convolutional layer to map $I_{LR}$ to an initial feature space:
\begin{equation}
    F_S = \text{Conv}(I_{LR}),
\end{equation}
where $F_S \in \mathbb{R}^{H \times W \times C}$ represents the shallow feature map, with $H$ and $W$ denoting the spatial dimensions of the input image, and $C$ the number of feature channels. This initial step provides a feature-rich representation for further processing.

\subsubsection{Deep Feature Extraction}

The deep feature extraction module builds upon the shallow features $F_S$ to capture complex spatial and channel relationships, outputting a deeper feature representation $F_D \in \mathbb{R}^{H \times W \times C}$. This module is composed of $N_1$ stacked Residual Groups (RGs), each containing $N_2$ Visual State Space (VSS) Blocks followed by an Overlapping Cross-Attention Block (OCAB). 

In each Residual Group (RG), features are processed through a sequence of VSS Blocks, arranged to optimize computational efficiency and performance in hybrid architectures. Specifically, we utilize a structure of six Mamba blocks followed by a single Transformer block, a ratio that has demonstrated effectiveness and was empirically validated in prior work \cite{waleffe2024empiricalstudymambabasedlanguage}. This setup aligns with the original HAT architecture, where we replaced the (S)W-MSA modules with Mamba blocks while preserving the original six-to-one block ratio. Although we considered adding more Transformer layers, the Overlapping Cross-Attention Block (OCAB) required for cross-window information flow is computationally intensive due to its overlapping nature. This increased the training time substantially and slowed the model to a point where further improvements in metrics could not justify the trade-off in efficiency, prompting us to maintain the initial six-to-one configuration.

The nested structure of the deep feature extraction module, composed of $N_1$ RGs, progressively refines the shallow features $F_S$. The output of each RG can be represented as:
\begin{equation}
    F_{RG}^{(j)} = \text{Conv}(\text{OCAB}(\text{VSS}_{N_2}(F_{RG}^{(j-1)}))) + F_{RG}^{(j-1)},
\end{equation}
where $j$ denotes the index of the current RG, $F_{RG}^{(j-1)}$ is the input to the RG, and $\text{VSS}_{N_2}$ represents the sequential application of $N_2$ VSS Blocks within the RG. Each VSS Block applies a state-space model to capture global dependencies with reduced computational overhead, as follows:
\begin{equation}
    F_{VSS}^{(i)} = \text{VSS}(F_{VSS}^{(i-1)}),
\end{equation}
where $i$ denotes the index of the VSS Block within an RG.

Following the VSS Blocks, the OCAB operation integrates cross-window information to enhance spatial coherence, ensuring a more continuous representation across feature windows. This is formulated as:
\begin{equation}
    F_{OCAB} = \text{OCAB}(F_{VSS}^{(N_2)}),
\end{equation}
where $F_{OCAB}$ represents the output after the OCAB step, which is then refined by a convolution layer to produce the final output of the RG.

Thus, after passing through all $N_1$ Residual Groups, the deep feature extraction module yields the deep feature representation $F_D$, where each RG applies the combined operations of VSS Blocks, OCAB, and residual connections to progressively enrich the feature map.

\subsubsection{Spatial Gated Feed-Forward Network (SGFN)}

\begin{figure}[t]
    \centering
    \begin{minipage}[t]{0.45\textwidth}
        \includegraphics[width=\textwidth]{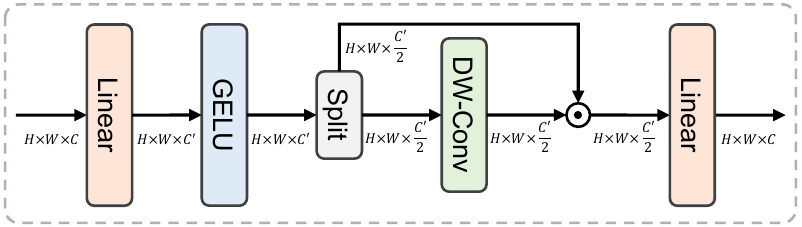}
        \caption{Illustration of spatial-gate feed-forward network.}
        \label{first_fig}
    \end{minipage}
\end{figure}

Instead of conventional MLP layers within each VSS Block, we employ the Spatial Gated Feed-Forward Network (SGFN) based on \cite{chen2023dual}. The SGFN adds a spatial gate to the Feed-Forward Network (FFN) layers, addressing the limitations of traditional FFNs in spatial modeling. The SGFN applies non-linear activation and two linear projections, with an additional spatial gate for selective channel-wise information flow. Overall, given the input $\hat{X}$$\in$$\mathbb{R}^{H \times W \times C}$, SGFN is formulated as
\begin{equation}
\begin{gathered}
\hat{X'} = \sigma(W_p^1 \hat{X}),\quad [\hat{X_1'}, \hat{X_2'}] = \hat{X'},\\
{\rm SGFN}(\hat{X}) = W_p^2(\hat{X_1'} \odot (W_d \hat{X_2'})),
\label{eq:sgfn}
\end{gathered}
\end{equation}
where $W_p^1$ and $W_p^2$ indicate linear projection layers, $\sigma$ is the GELU activation function, and $W_d$ represents the learnable parameters for the depth-wise convolution. Both $\hat{X_1'}$ and $\hat{X_2'}$ lie in $\mathbb{R}^{H \times W \times \frac{C'}{2}}$ space, with $C'$ denoting SGFN's hidden dimension. Unlike the standard FFN, SGFN is designed to capture non-linear spatial information and reduce channel redundancy in fully-connected layers.

\subsubsection{Image Reconstruction}

The image reconstruction module takes the deep feature representation $F_D$ and upscales it to produce the final high-resolution (HR) output image $I_{HR} \in \mathbb{R}^{H_{out} \times W_{out} \times 3}$. The upsampling is performed using the pixel shuffle method \cite{shi2016real}, ensuring efficient spatial upscaling:
\begin{equation}
    F_{up} = \text{PixelShuffle}(F_D),
\end{equation}
where $F_{up}$ is the upscaled feature map. After upsampling, a convolutional layer aggregates the upscaled features to produce the HR image:
\begin{equation}
    I_{HR} = \text{Conv}(F_{up}).
\end{equation}

This series of transformations enhances the input LR image into a high-quality SR output, maintaining spatial continuity and ensuring fine-grained detail preservation across the reconstructed image.

\section{Experiments}

\subsection{Experimental Settings}

\textbf{Implementation Details.} We build two variants of the Contrast model with different complexity levels, called Contrast and Contrast-S. For the main Contrast model, we use 6 Residual Groups (RGs) with an embedding dimension of 210, a window size of 32, an MLP ratio of 2, an SSM state dimension of 1, and an SSM ratio of 1, as shown in the VMamba architecture. In the Contrast-S variant, we also use 6 RGs but with a reduced embedding dimension of 150 and a smaller window size of 16, while keeping the MLP ratio, SSM state dimension, and SSM ratio the same as in the main model.

\textbf{Data and Evaluation.} We follow the standard practices established in previous works \cite{chen2022cross, chen2023dual} for training and evaluating our models. Specifically, we use two large-scale datasets for training: DIV2K \cite{timofte2017ntire} and Flickr2K \cite{lim2017enhanceddeepresidualnetworks}. For evaluation, we test on five widely-used benchmark datasets: Set5 \cite{bevilacqua2012low}, Set14 \cite{zeyde2010single}, B100 \cite{martin2001database}, Urban100 \cite{huang2015single}, and Manga109 \cite{matsui2017sketch}. 

Our experiments cover upscaling factors of $\times$2, $\times$3, and $\times$4. Low-resolution (LR) images are generated from high-resolution (HR) images using bicubic degradation. To evaluate super-resolution (SR) performance, we use Peak Signal-to-Noise Ratio (PSNR) and Structural Similarity Index (SSIM) \cite{wang2004image}, both calculated on the Y channel (luminance) of the YCbCr color space. These metrics provide objective assessments of image quality and structural preservation in SR results.

\textbf{Training Settings.} We train the models with a patch size of 64×64 and a batch size of 32 for 500K iterations. The optimization is performed by minimizing the L1 loss using the Adam optimizer \cite{kingma2014adam} with $\beta_1 = 0.9$ and $\beta_2 = 0.99$. The initial learning rate is set to $1 \times 10^{-4}$ for the main Contrast model and $2 \times 10^{-4}$ for Contrast-S. The learning rate is halved at milestones: [250K, 400K, 450K, 475K]. During training, data augmentation is applied by randomly rotating images by 90°, 180°, and 270°, along with horizontal flips. Our model is implemented in PyTorch\cite{paszke2019pytorch} and Triton\cite{tillet2019triton} and trained on 4 A6000 GPUs.

\subsection{Ablation Study}

In this section, we conduct a series of ablation studies to investigate the effectiveness of different components in our proposed \textbf{Contrast} model. We aim to understand how the integration of Transformer and State Space components contributes to the model's performance, and we analyze the impact of architectural choices on training dynamics and final results. All experiments are conducted with models of comparable parameter sizes to ensure a fair comparison.

\textbf{Effect of Replacing Transformer Blocks with SS2D.} We first examine the impact of replacing the (Shifted) Window Multi-Head Self-Attention ((S)W-MSA) blocks in the baseline HAT~\cite{chen2023activating} with the State Space 2D (SS2D) blocks from Mamba~\cite{mamba}. This yields an initial hybrid design, allowing us to assess whether SS2D can capture dependencies and spatial context more effectively than conventional self-attention mechanisms.

\textbf{Training Dynamics of Pure Mamba, Transformer, and Hybrid Models.} To further analyze the individual contributions of Mamba and Transformer architectures, we train three models:

\begin{itemize} \item \textbf{Pure Mamba:} All Transformer blocks are replaced with SS2D blocks, and Channel Attention Blocks (CABs) are removed. \item \textbf{Pure Transformer:} The baseline HAT model without modifications. \item \textbf{Hybrid Contrast:} The baseline HAT model in which (S)W-MSA blocks are replaced by SS2D blocks. \end{itemize}

\begin{figure}[th] \centering \includegraphics[width=0.50\textwidth]{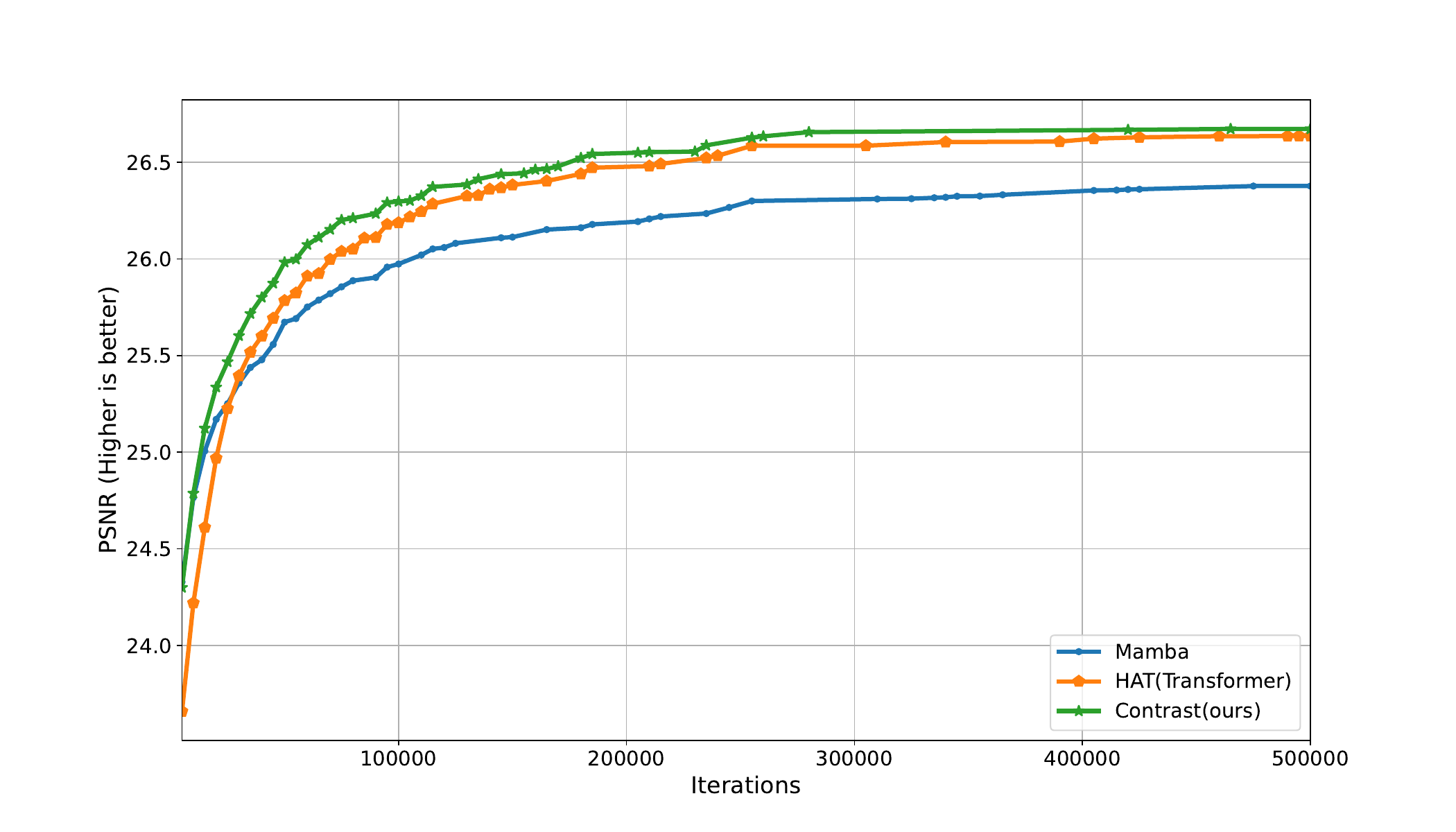} \vspace{-5mm} \caption{PSNR (dB) during training of pure Mamba~\cite{mamba}, pure Transformer (HAT~\cite{chen2023activating}), and our proposed Hybrid Contrast model. Evaluated on the Urban100~\cite{huang2015single} dataset for $\times4$ SR.} \label{fig:training_dynamics} \end{figure}

All three models contain a comparable number of parameters for a fair comparison. Figure~\ref{fig:training_dynamics} illustrates an interesting phenomenon: \emph{Pure Mamba} quickly attains high PSNR early in training but plateaus over time, whereas the \emph{Transformer} model begins with lower PSNR yet continually improves, eventually surpassing Mamba. The \emph{Hybrid Contrast} approach leverages both advantages: it starts strong—akin to Mamba—and continues to improve consistently—similar to the Transformer—leading to superior final performance.

This behavior remains consistent across Set5, Set14, B100, Urban100, and Manga109. The hybrid model effectively merges Mamba’s rapid early learning with the Transformer’s sustained improvement, demonstrating the complementarity of SS2D blocks and Transformer layers in super-resolution tasks.

\textbf{Ratio of Mamba to Transformer Blocks.} In HAT~\cite{chen2023activating}, (S)W-MSA is combined with Overlapping Cross-Attention (OCA) at a ratio of 6:1. Meanwhile, recent work~\cite{waleffe2024empiricalstudymambabasedlanguage} has shown that a similar ratio of Mamba to Transformer blocks can be particularly efficient. Inspired by these findings, our hybrid approach replaces (S)W-MSA with Mamba blocks while preserving this ratio, thereby minimizing architectural changes and training overhead. We experimented with alternative configurations: reducing the number of Mamba blocks demanded more Transformer blocks (and hence more OCA), which substantially slowed training due to OCA’s higher computational cost. Conversely, increasing the number of Mamba blocks further yielded diminishing benefits from the Transformer layers, leading to minimal performance gains. Adhering to this established 6:1 ratio thus strikes an effective balance between accuracy and training efficiency in our hybrid Contrast model.

\textbf{Impact of Channel Attention Blocks.} We also investigated the role of Channel Attention Blocks (CABs) in our hybrid model. CABs were originally included in the baseline, but we wanted to evaluate whether they are essential to our hybrid architecture or if their removal might yield a more efficient design without degrading performance. To this end, we compared two versions of the hybrid model:

\begin{itemize}
    \item \textbf{With CAB:} The hybrid model including CABs.
    \item \textbf{Without CAB:} The hybrid model without CABs, featuring extra layers and adjusted embedding dimensions to maintain roughly the same number of parameters.
\end{itemize}

Table~\ref{tab:ablation_cab} shows that removing CABs not only simplifies the model but also delivers higher PSNR values on all examined datasets. Based on these findings, we decided to omit CABs in our final model, as the hybrid architecture benefits more from deeper layers and increased embedding dimensions than from channel attention.

\begin{table}[ht]
\centering
\small
\setlength{\tabcolsep}{4pt}
\caption{Ablation study on the impact of Channel Attention Blocks (CABs). We compare models with and without CABs, adjusting the number of layers and embedding dimensions to keep the parameter count approximately equal. Results are reported in PSNR (dB) for $\times$4 SR.}
\begin{tabular}{ l c c c c c }
\hline
Model & Set5 & Set14 & B100 & Urban100 & Manga109 \\ \hline
With CAB & 32.44 & 28.78 & 27.68 & 26.57 & 31.06 \\
Without CAB & \textbf{32.53} & \textbf{28.88} & \textbf{27.69} & \textbf{26.65} & \textbf{31.16} \\ \hline
\end{tabular}
\label{tab:ablation_cab}
\end{table}

\textbf{Replacing MLP with SGFN.} We further explored the effect of replacing the standard Multi-Layer Perceptron (MLP) layers in our model with the Spatial Gated Feed-forward Network (SGFN). The SGFN is designed to enhance the model's ability to capture spatial information by integrating gating mechanisms. We trained two versions of the model:

\begin{itemize}
    \item With MLP: The hybrid model using standard MLP layers.
    \item With SGFN: The hybrid model with MLP layers replaced by SGFN.
\end{itemize}

The results, shown in Table~\ref{tab:ablation_sgfn}, demonstrate that using SGFN leads to improved performance on most datasets.

\begin{table}[ht]
\centering
\small % Reduces the font size of the table
\setlength{\tabcolsep}{4pt} % Adjusts the space between columns
\caption{Comparison of models using MLP and SGFN layers. Results are reported in PSNR (dB) for $\times$4 SR.}
\begin{tabular}{ l c c c c c }
\hline
Model & Set5 & Set14 & B100 & Urban100 & Manga109 \\ \hline
With MLP & 32.44 & 28.84 & 27.73 & 26.72 & 31.22 \\
With SGFN & \textbf{32.50} & \textbf{28.87} & 27.73 & \textbf{26.74} & \textbf{31.24} \\ \hline
\end{tabular}
\label{tab:ablation_sgfn}
\end{table}

\begin{table*}[t]
\scriptsize
\begin{center}
\begin{tabular}{l|c|cccccccccccccccc} 
\toprule[0.15em]
	  &  &  \multicolumn{2}{c}{Set5} & \multicolumn{2}{c}{Set14} & \multicolumn{2}{c}{B100} & \multicolumn{2}{c}{Urban100} & \multicolumn{2}{c}{Manga109}\\

	 \multirow{-2}{*}{Method}& \multirow{-2}{*}{Scale} & PSNR & SSIM & PSNR & SSIM & PSNR & SSIM & PSNR & SSIM & PSNR & SSIM
\\
\midrule[0.15em]
EDSR~\cite{lim2017enhanced} & $\times$2 & 
38.11 & 0.9602 & 33.92 & 0.9195 & 32.32 & 0.9013 & 32.93 & 0.9351 & 39.10 & 0.9773\\
RCAN~\cite{zhang2018image} & $\times$2 &
38.27 & 0.9614 & 34.12 & 0.9216 & 32.41 & 0.9027 & 33.34 & 0.9384 & 39.44 & 0.9786\\
SAN~\cite{Dai_2019_CVPR} & $\times$2 &
38.31 & 0.9620 & 34.07 & 0.9213 & 32.42 & 0.9028 & 33.10 & 0.9370 & 39.32 & 0.9792\\
HAN~\cite{niu2020singleimagesuperresolutionholistic} & $\times$2 &
38.27 & 0.9614 & 34.16 & 0.9217 & 32.41 & 0.9027 & 33.35 & 0.9385 & 39.46 & 0.9785\\
CSNLN~\cite{mei2020imagesuperresolutioncrossscalenonlocal} & $\times$2 &
38.28 & 0.9616 & 34.12 & 0.9223 & 32.40 & 0.9024 & 33.25 & 0.9386 & 39.37 & 0.9785\\
NLSA~\cite{mei2021imageNLSA} & $\times$2 &
38.34 & 0.9618 & 34.08 & 0.9231 & 32.43 & 0.9027 & 33.42 & 0.9394 & 39.59 & 0.9789\\
ELAN~\cite{zhang2022efficient} & $\times$2 &
38.36 & 0.9620 & 34.20 & 0.9228 & 32.45 & 0.9030 & 33.44 & 0.9391 & 39.62 & 0.9793\\
DFSA~\cite{abdelmagid2021dynamic} & $\times$2 &
38.38 & 0.9620 & 34.33 & 0.9232 & 32.50 & 0.9036 & 33.66 & 0.9412 & 39.98 & 0.9798\\
SwinIR~\cite{liang2021swinir} & $\times$2 &  
{38.42} & {0.9623} & {34.46} & {0.9250} & {32.53} &{0.9041} & {33.81} &{0.9427} & {39.92} & {0.9797}\\
Swin2SR~\cite{conde2022swin2sr} & $\times$2 &  
{38.43} & {0.9623} & {34.48} & {0.9256} & {32.54} &{0.9050} & {33.89} &{0.9431} & {39.88} & {0.9798}\\
ART~\cite{zhang2023accurateimagerestorationattention} & $\times$2 &  
{38.56} & {0.9629} & {34.59} & \textcolor{blue}{0.9267} & {32.58} &{0.9048} & {34.30} &{0.9452} & {40.24} & {0.9808}\\
CAT-A~\cite{chen2022cross} & $\times$2 &  
38.51 & 0.9626 & \textcolor{blue}{34.78} & 0.9265 & 32.59 & 0.9047 & 34.26 & 0.9440 & 40.10 & 0.9805
\\ 
SRFormer~\cite{zhou2023srformer} & $\times$2 &  
{38.51} & {0.9627} & {34.44} & {0.9253} & {32.57} &{0.9046} & {34.09} &{0.9449} & {40.07} & {0.9802}\\
HAT~\cite{chen2023activating} & $\times$2 &  
\textcolor{red}{38.63} & \textcolor{blue}{0.9630} & \textcolor{red}{34.86} & \textcolor{red}{0.9274} &\textcolor{red}{32.62} &\textcolor{red}{0.9053} & \textcolor{blue}{34.45} &\textcolor{blue}{0.9466} & {40.26} & \textcolor{blue}{0.9809}\\
MambaIR~\cite{guo2024mambair} & $\times$2 &  
{38.57} & {0.9627} & {34.67} & {0.9261} &{32.58} &{0.9048} & {34.15} &{0.9446} & \textcolor{blue}{40.28} & {0.9806}\\
Contrast-S (ours) & $\times$2 &  
{38.49} & {0.9624} & {34.58} & {0.9256} &{32.54} &{0.9041} & {34.06} &{0.9438} & {40.06} & {0.9802}\\
Contrast (ours) & $\times$2 &  
\textcolor{blue}{38.58} & \textcolor{red}{0.9630} & {34.68} & {0.9261} &\textcolor{blue}{32.59} &\textcolor{blue}{0.9050} & \textcolor{red}{34.45} &\textcolor{red}{0.9467} & \textcolor{red}{40.30} & \textcolor{red}{0.9809}\\

\midrule
EDSR~\cite{lim2017enhanced} & $\times$3 & 
34.65 & 0.9280 & 30.52 & 0.8462 & 29.25 & 0.8093 & 28.80 & 0.8653 & 34.17 & 0.9476\\
RCAN~\cite{zhang2018image} & $\times$3 &
34.74 & 0.9299 & 30.65 & 0.8482 & 29.32 & 0.8111 & 29.09 & 0.8702 & 34.44 & 0.9499\\
SAN~\cite{Dai_2019_CVPR} & $\times$3 &
34.75 & 0.9300 & 30.59 & 0.8476 & 29.33 & 0.8112 & 28.93 & 0.8671 & 34.30 & 0.9494\\
HAN~\cite{niu2020singleimagesuperresolutionholistic} & $\times$3 &
34.75 & 0.9299 & 30.67 & 0.8483 & 29.32 & 0.8110 & 29.10 & 0.8705 & 34.48 & 0.9500\\
CSNLN~\cite{mei2020imagesuperresolutioncrossscalenonlocal} & $\times$3 &
34.74 & 0.9300 & 30.66 & 0.8482 & 29.33 & 0.8105 & 29.13 & 0.8712 & 34.45 & 0.9502\\
NLSA~\cite{mei2021imageNLSA} & $\times$3 &
34.85 & 0.9306 & 30.70 & 0.8485 & 29.34 & 0.8117 & 29.25 & 0.8726 & 34.57 & 0.9508\\
ELAN~\cite{zhang2022efficient} & $\times$3 &
34.90 & 0.9313 & 30.80 & 0.8504 & 29.38 & 0.8124 & 29.32 & 0.8745 & 34.73 & 0.9517\\
DFSA~\cite{abdelmagid2021dynamic} & $\times$3 &
34.92 & 0.9312 & 30.83 & 0.8507 & 29.42 & 0.8128 & 29.44 & 0.8761 & 35.07 & 0.9525\\
SwinIR~\cite{liang2021swinir} & $\times$3 &  
{34.97} & {0.9318} & {30.93} & {0.8534} & {29.46} & {0.8145} & {29.75} & {0.8826} & {35.12} & {0.9537}
\\
ART~\cite{zhang2023accurateimagerestorationattention} & $\times$3 &  
{35.07} & {0.9325} & {31.02} & {0.8541} & {29.51} & {0.8159} & {30.10} & {0.8871} & {35.39} & {0.9548}
\\
CAT-A~\cite{chen2022cross} & $\times$3 &  
35.06 & \textcolor{blue}{0.9326} & \textcolor{blue}{31.04} & 0.8538 & 29.52 & 0.8160 & 30.12 & 0.8862 & 35.38 & 0.9546
\\
SRFormer~\cite{zhou2023srformer} & $\times$3 &  
{35.02} & {0.9323} & {30.94} & {0.8540} & {29.48} & {0.8156} & {30.04} & {0.8865} & {35.26} & {0.9543}
\\
HAT~\cite{chen2023activating} & $\times$3 &  
\textcolor{blue}{35.07} & \textcolor{red}{0.9329} & \textcolor{red}{31.08} & \textcolor{red}{0.8555} & \textcolor{red}{29.54} & \textcolor{red}{0.8167} & \textcolor{red}{30.23} & \textcolor{red}{0.8896} & \textcolor{red}{35.53} & \textcolor{red}{0.9552}
\\
MambaIR~\cite{guo2024mambair} & $\times$3 &  
\textcolor{red}{35.08} & {0.9323} & {30.99} & {0.8536} & {29.51} & {0.8157} & {29.93} & {0.8841} & {35.43} & {0.9546}
\\
Contrast-S (ours) & $\times$3 &  
{35.02} & {0.9322} & {30.97} & {0.8541} & {29.49} & {0.8151} & {29.93} & {0.8838} & {35.25} & {0.9538}
\\
Contrast (ours) & $\times$3 &  
{35.06} & {0.9324} & {31.00} & \textcolor{blue}{0.8541} & \textcolor{blue}{29.52} & \textcolor{blue}{0.8160} & \textcolor{blue}{30.17} & \textcolor{blue}{0.8884} & \textcolor{blue}{35.45} & \textcolor{blue}{0.9549}
\\

\midrule
EDSR~\cite{lim2017enhanced} & $\times$4 & 
32.46 & 0.8968 & 28.80 & 0.7876 & 27.71 & 0.7420 & 26.64 & 0.8033 & 31.02 & 0.9148\\
RCAN~\cite{zhang2018image} & $\times$4 &
32.63 & 0.9002 & 28.87 & 0.7889 & 27.77 & 0.7436 & 26.82 & 0.8087 & 31.22 & 0.9173\\
SAN~\cite{Dai_2019_CVPR} & $\times$4 &
32.64 & 0.9003 & 28.92 & 0.7888 & 27.78 & 0.7436 & 26.79 & 0.8068 & 31.18 & 0.9169\\
HAN~\cite{niu2020singleimagesuperresolutionholistic} & $\times$4 &
32.64 & 0.9002 & 28.90 & 0.7890 & 27.80 & 0.7442 & 26.85 & 0.8094 & 31.42 & 0.9177\\
CSNLN~\cite{mei2020imagesuperresolutioncrossscalenonlocal} & $\times$4 &
32.68 & 0.9004 & 28.95 & 0.7888 & 27.80 & 0.7439 & 27.22 & 0.8168 & 31.43 & 0.9201\\
NLSA~\cite{mei2021imageNLSA} & $\times$4 &
32.59 & 0.9000 & 28.87 & 0.7891 & 27.78 & 0.7444 & 26.96 & 0.8109 & 31.27 & 0.9184\\
ELAN~\cite{zhang2022efficient} & $\times$4 &
32.75 & 0.9022 & 28.96 & 0.7914 & 27.83 & 0.7459 & 27.13 & 0.8167 & 31.68 & 0.9226\\
DFSA~\cite{abdelmagid2021dynamic} & $\times$4 &
32.79 & 0.9019 & 29.06 & 0.7922 & 27.87 & 0.7458 & 27.17 & 0.8163 & 31.88 & 0.9266\\
SwinIR~\cite{liang2021swinir} & $\times$4 & 
{32.92} & {0.9044} & {29.09} & {0.7950} & {27.92} & {0.7489} & {27.45} & {0.8254} & {32.03} & {0.9260}\\
Swin2SR~\cite{conde2022swin2sr} & $\times$4 & 
32.92 & 0.9039 & 29.06 & 0.7946 & 27.92 & 0.7505 & 27.51 & 0.8271 & 31.03 & 0.9256
\\
ART~\cite{zhang2023accurateimagerestorationattention} & $\times$4 & 
{33.04} & 0.9051 & 29.16 & 0.7958 & 27.97 & 0.7510 & 27.77 & 0.8321 & 32.31 & 0.9283
\\
CAT-A~\cite{chen2022cross} & $\times$4 & 
33.08 & 0.9052 & 29.18 & 0.7960 & 27.99 & 0.7510 & 27.89 & 0.8339 & 32.39 & 0.9285
\\
SRFormer~\cite{zhou2023srformer} & $\times$4 & 32.93 & 0.9041 & 29.08 & 0.7953 & 27.94 & 0.7502 & 27.68 & 0.8311 & 32.21 & 0.9271
\\
HAT~\cite{chen2023activating} & $\times$4 & \textcolor{blue}{33.04} & \textcolor{red}{0.9056} & \textcolor{blue}{29.23} & 0.7973 & \textcolor{blue}{28.00} & \textcolor{blue}{0.7517} & \textcolor{blue}{27.97} & \textcolor{blue}{0.8368} & \textcolor{blue}{32.48} & \textcolor{blue}{0.9292}
\\
MambaIR~\cite{guo2024mambair} & $\times$4 & 33.03 & 0.9046 & 29.20 & 0.7961 & 27.98 & 0.7503 & 27.68 & 0.8287 & 32.32 & 0.9272
\\
Contrast-S (ours) & $\times$4 & 32.92 & 0.9038 & 29.11 & 0.7947 & 27.93 & 0.7489 & 27.59 & 0.8280 & 32.21 & 0.9264
\\
Contrast (ours) & $\times$4 & 32.94 & 0.9030 & 29.20 & \textcolor{blue}{0.7973} & 27.98 & 0.7508 & 27.92 & 0.8357 & 32.38 & 0.9283
\\
Contrast+ (ours) & $\times$4 & \textcolor{red}{33.09} & \textcolor{blue}{0.9055} & \textcolor{red}{29.23} & \textcolor{red}{0.7980} & \textcolor{red}{28.00} & \textcolor{red}{0.7520} & \textcolor{red}{28.07} & \textcolor{red}{0.8385} & \textcolor{red}{32.61} & \textcolor{red}{0.9299}
\\
\bottomrule[0.15em]
\end{tabular}
\vspace{-1mm}
\caption{\small{Quantitative comparison with state-of-the-art methods. The best and second-best results are coloured \textcolor{red}{red} and \textcolor{blue}{blue}.}}
\label{table:comparison_tab}
\end{center}
\vspace{-8mm}
\end{table*}

\begin{figure}[th]
    \centering
    \includegraphics[width=0.45\textwidth]{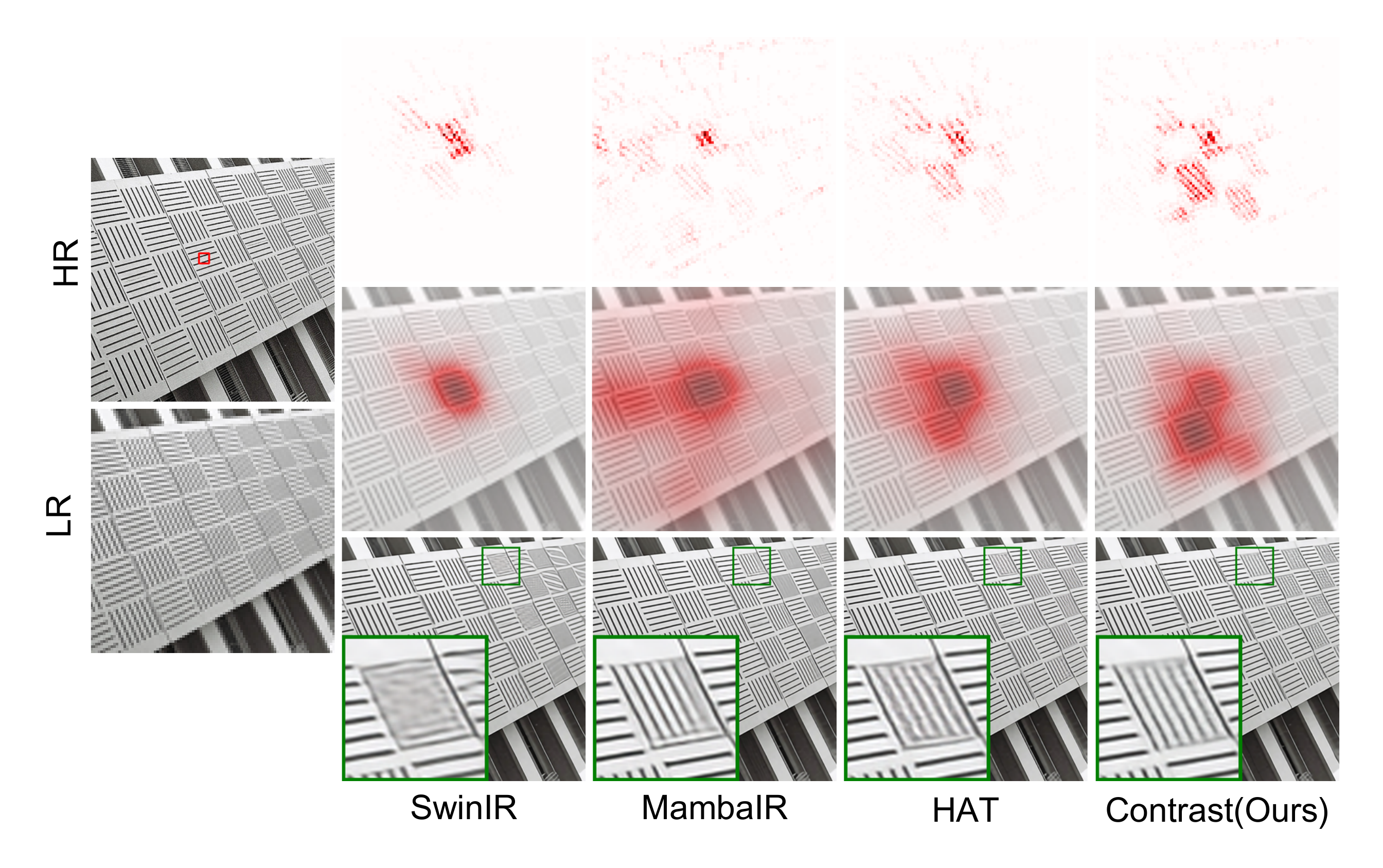}
    \caption{LAM visualizations and prediction results. The first and second rows show LAM visualizations for MambaIR~\cite{guo2024mambair}, the Transformer model, and our hybrid Contrast model. The third row presents the corresponding SR predictions. The red square on the HR image marks the region used for LAM calculation. MambaIR exhibits a more global receptive field with a preference for features aligned horizontally and vertically, while the Transformer focuses more locally. Our Contrast model effectively combines these characteristics, capturing both global context and precise spatial details.}
    \label{fig:lam_visualization}
\end{figure}

\textbf{LAM Visualization Analysis.} To better understand how each model captures dependencies and spatial context, we utilize local attribution map (LAMs)~\cite{gu2021interpreting}. Figure~\ref{fig:lam_visualization} presents the LAM visualizations for pure Mamba, pure Transformer, and our hybrid Contrast model.

From the LAMs, we observe that MambaIR has a global receptive field, but with a tendency to emphasize horizontal and vertical features over diagonal ones. The Transformer model shows a more localized attention pattern. Our hybrid Contrast model successfully captures a broad receptive field while maintaining focus on important spatial regions, especially those with higher resolution or closer to the camera. This visualization underscores the advantage of our hybrid approach in balancing global context modeling and precise spatial attention, supporting our quantitative findings.

\begin{table}[ht]
\centering
\small
\setlength{\tabcolsep}{4pt}
\caption{Model complexity comparisons ($\times$4). PSNR (dB) on Urban100 and Manga109, FLOPs, and Params are reported. FLOPs are measured with a 3\(\times\)256\(\times\)256 output.}
\begin{tabular}{lccccc}
\hline
Method & HAT & MambaIR & Contrast-S & Contrast \\ \hline
Params (M) & 20.77 & 20.57 & 7.55 & 14.29 \\
FLOPs (G) & 172.06 & 140.26 & 61.44 & 113.74 \\
Urban100($\times2$) & 34.45 & 34.15 & 34.06 & 34.45 \\
Manga109($\times2$) & 40.26 & 40.28 & 40.06 & 40.30 \\\hline
\end{tabular}
\label{tab:model_size}
\end{table}

\begin{figure*}[ht]
    \centering
    \includegraphics[width=\textwidth]{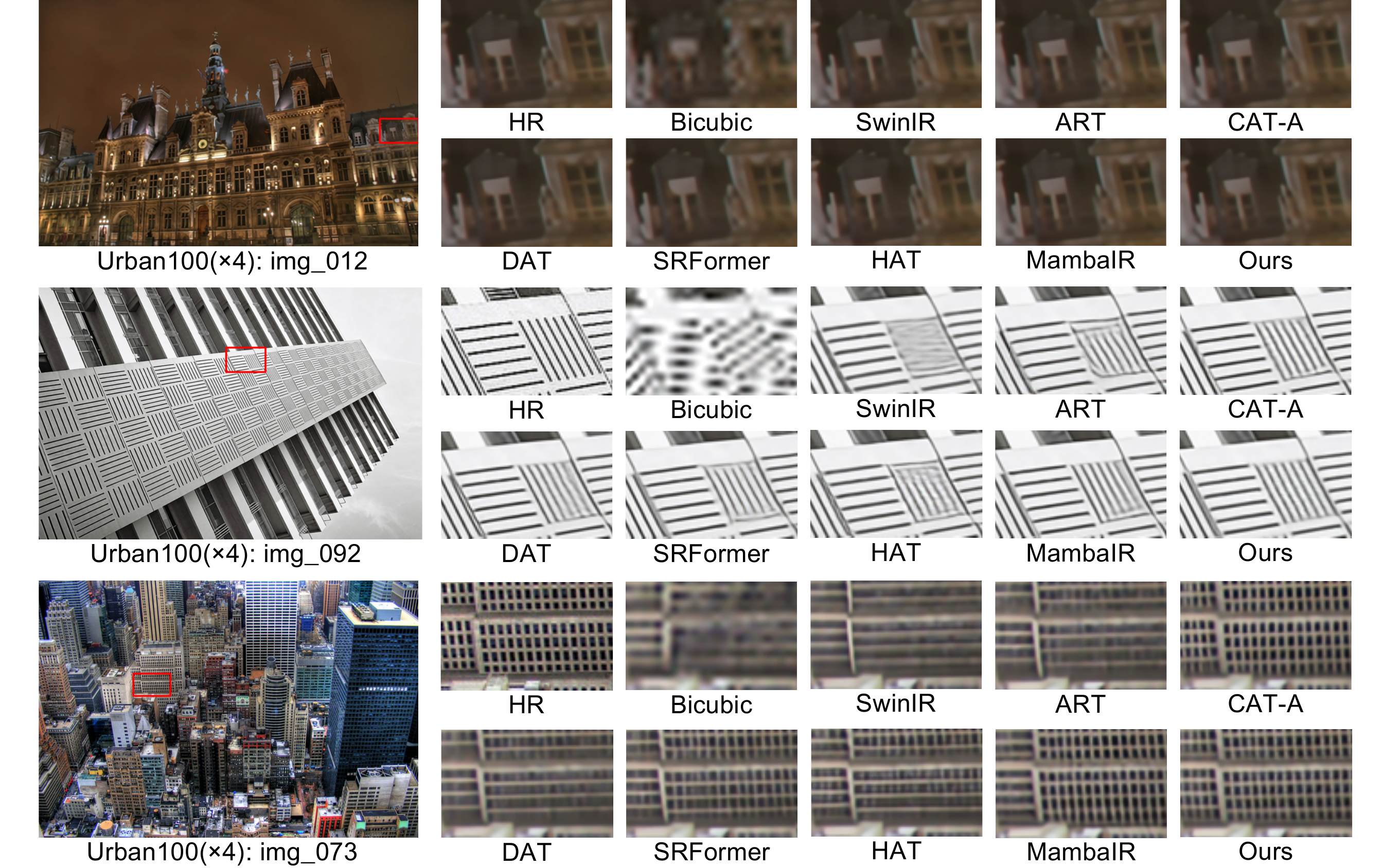}
    \caption{Visual comparison for $\times$4 SR on the Urban100 dataset. Our Contrast model produces sharper and more detailed images, effectively reconstructing fine textures and patterns.}
    \label{fig:visual_comparison}
\end{figure*}

\subsection{Comparison with State-of-the-Art Methods}

\textbf{Model Size Analysis.} In Table~\ref{tab:model_size}, we compare the complexity of various methods at $\times4$ scale, where FLOPs are calculated for a 3\(\times\)256\(\times\)256 output. Notably, both \textbf{Contrast-S} (7.55M) and \textbf{Contrast} (14.29M) use significantly fewer parameters compared to HAT (20.77M) and MambaIR (20.57M), yet achieve competitive or better performance on Urban100$\times2$ and Manga109$\times2$. For instance, \textbf{Contrast} surpasses MambaIR by 0.30~dB on Urban100 and matches HAT on Urban100, all while using fewer parameters. These results highlight our model’s efficiency in balancing computational cost and SR accuracy, particularly when compared to leading methods in Transformer-based (HAT) and state space-based (Mamba) architectures.

\textbf{Quantitative Results.} We further compare our methods (\textbf{Contrast-S} and \textbf{Contrast}) with leading SR approaches in Table~\ref{table:comparison_tab}. Both models deliver strong performance, particularly on Urban100~\cite{huang2015single}, which contains complex structures and repetitive patterns. \textbf{Contrast} attains results comparable to HAT~\cite{chen2023activating} while employing noticeably fewer parameters, and it also clearly outperforms MambaIR~\cite{guo2024mambair}. Our smaller variant, \textbf{Contrast-S}, remains highly competitive despite having less than half the parameters of MambaIR, further demonstrating the efficiency gains of our hybrid design.

\textbf{Visual Results.} Figure~\ref{fig:visual_comparison} shows examples from the Urban100 dataset. Thanks to the combined strengths of Mamba’s broad receptive field and the Transformer’s precise local context modeling, \textbf{Contrast} reconstructs sharper details and clearer textures compared to other state-of-the-art methods. This highlights the practical effectiveness of our approach for high-resolution SR tasks requiring both global context and fine-grained accuracy.

\section{Conclusion}

In this paper, we presented \textbf{Contrast}, a novel hybrid super-resolution model that combines convolutional layers, Transformer blocks, and state space components in a unified framework. By blending Mamba’s global receptive field with the precise attention capabilities of Transformers, our method addresses the respective limitations of each approach—namely, the high computational overhead of pure Transformers in handling global context and the approximate nature of Mamba’s hidden state mechanisms at the pixel level. Additionally, we incorporated SGFNs (convolution-based modules) from prior work in place of MLP blocks to enhance local dependencies.

Our experimental results highlight \textbf{Contrast}’s strong performance on challenging datasets such as Urban100 and Manga109, which both contain high-resolution images with intricate structures. This demonstrates the model’s effectiveness in simultaneously capturing broad global context and fine-grained details. In summary, \textbf{Contrast} underscores the promise of hybrid architectures in advancing state-of-the-art super-resolution, paving the way for future research that balances efficiency and high-quality image reconstruction in low-level computer vision.

\noindent \textbf{Future Work.} Larger model capacities promise to further leverage Mamba’s broad receptive field and the Transformer’s precise local attention, potentially boosting overall SR performance. Moving beyond classical super-resolution, \textbf{Contrast} could be adapted to additional low-level vision tasks such as real-world super-resolution, JPEG compression artifact reduction, and both color and grayscale denoising, extending its efficacy to more diverse domains. Furthermore, exploring multi-scale or domain-specific training protocols may refine its adaptability to real-world deployment settings where computational resources and data characteristics vary significantly.
{
    \small
    \bibliographystyle{ieeenat_fullname}
    \bibliography{main}
}


\end{document}